\documentclass[letterpaper, 10 pt, journal]{IEEEtran}  

\IEEEoverridecommandlockouts                              

\usepackage{amsmath} 
\usepackage{amsthm}
\usepackage{amssymb}  
\usepackage{lipsum}
\usepackage{tikz}
\usetikzlibrary{shapes.geometric}
\usepackage{bbm}

\usepackage[english]{babel}

\usepackage[backend=biber, style=numeric, maxnames=1, giveninits=true, sorting=none, doi=false,isbn=false,url=false, eprint=false]{biblatex}
\usepackage{algorithm}
\usepackage{algpseudocode}

\usepackage{caption}
\usepackage{subcaption}
\usepackage[T1]{fontenc}
\usepackage{dblfloatfix}
\usepackage{hyperref}

\usepackage{tikz}
\usepackage{pifont}

\usepackage{url}

\newcommand{\bt}{\mathcal{T}}
\title{\LARGE \bf Progress Constraints for Reinforcement Learning in Behavior Trees}

\author{Finn Rietz$^{1, \dag}$, Mart Kartašev$^{2, \dag}$, Petter Ögren$^{2}$,  Johannes A. Stork$^{1}$%
\thanks{$^{\dag}$Equal contribution}%
\thanks{$^{1}$Adaptive and Interpretable Learning Systems Lab, Department of Computer Science, Örebro University.}%
\thanks{$^{2}$Robotics, Perception and Learning Lab, School of Electrical Engineering and Computer Science, Royal Institute of Technology.}%
\thanks{This work was partially supported by the Wallenberg AI, Autonomous Systems and Software Program (WASP) funded by the Knut and Alice Wallenberg Foundation.}%
}

\usepackage{changes}

\presetkeys%
{todonotes}{inline,backgroundcolor=yellow}{}

\newtheorem{definition}{Definition}

\newtheorem{remark}{Remark}
\newtheorem{theorem}{Theorem}

\pdfoutput=1
\usepackage[T1]{fontenc}

\begin{document}

\maketitle
\thispagestyle{empty}
\pagestyle{empty}

\begin{abstract}
Behavior Trees (BTs) provide a structured and reactive framework for decision-making, commonly used to switch between sub-controllers based on environmental conditions. 
Reinforcement Learning (RL), on the other hand, can learn near-optimal controllers but sometimes struggles with sparse rewards, safe exploration, and long-horizon credit assignment.
Combining BTs with RL has the potential for mutual benefit: a BT design encodes structured domain knowledge that can simplify RL training, while RL enables automatic learning of the controllers within BTs.
However, na\"ive integration of BTs and RL can lead to some controllers counteracting other controllers, possibly undoing previously achieved subgoals, thereby degrading the overall performance.
To address this, we propose \textit{progress constraints}, a novel mechanism where feasibility estimators constrain the allowed action set based on theoretical BT convergence results.
Empirical evaluations in a 2D proof-of-concept and a high-fidelity warehouse environment demonstrate improved performance, sample efficiency, and constraint satisfaction, compared to prior methods of BT-RL integration.
\end{abstract}

\section{INTRODUCTION AND MOTIVATION}
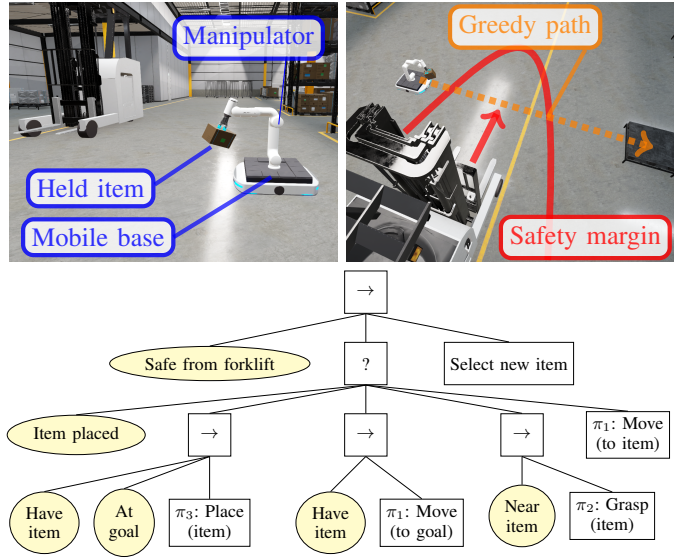
\begin{figure}
    \centering

    \begin{subfigure}[t]{0.24\textwidth}
                \begin{tikzpicture}
            \node at (0,0) {\includegraphics[width=\linewidth]{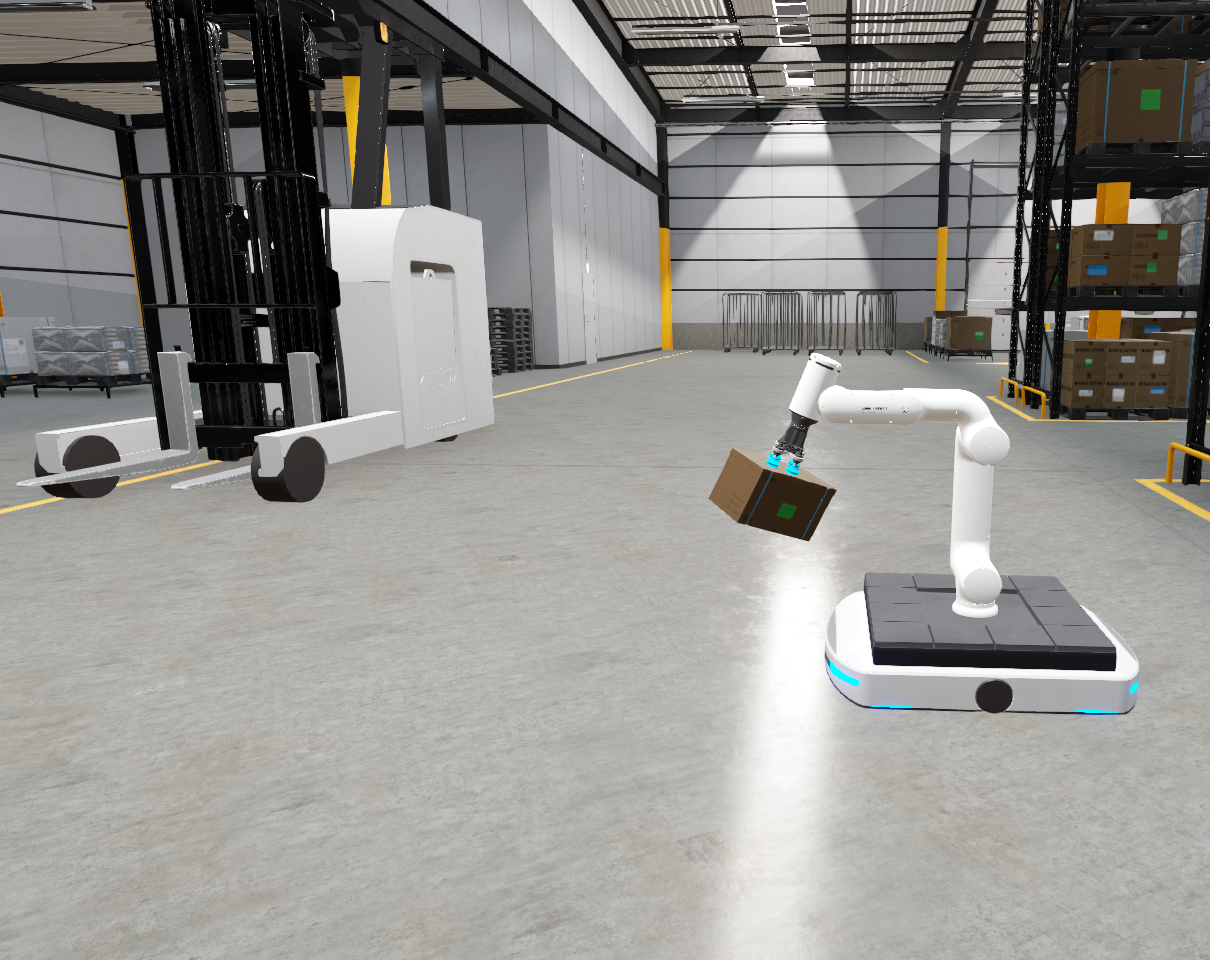}};
            \node[draw, blue, ultra thick, rounded corners, fill=white, opacity=0.8] at (1,1.25) {Manipulator};
            \draw[blue] (1.4, 0.1) -- (1.5, 1);
            
            \node[draw, blue,ultra thick, rounded corners, fill=white, opacity=0.8] at (-1,-1.4) {Mobile base};
            \draw[blue, ultra thick, opacity=0.7] (1.3, -0.6) -- (-0., -1.4);

            \node[draw, blue, ultra thick, rounded corners, fill=white, opacity=0.8] at (-1.1,-0.75) {Held item};
            \draw[blue, ultra thick, opacity=0.7] (0.5, -0.2) -- (-0.25, -0.7);

        \end{tikzpicture}
    \end{subfigure}
    \begin{subfigure}[t]{0.24\textwidth}
        \begin{tikzpicture}
            \node at (0,0) {\includegraphics[width=\linewidth]{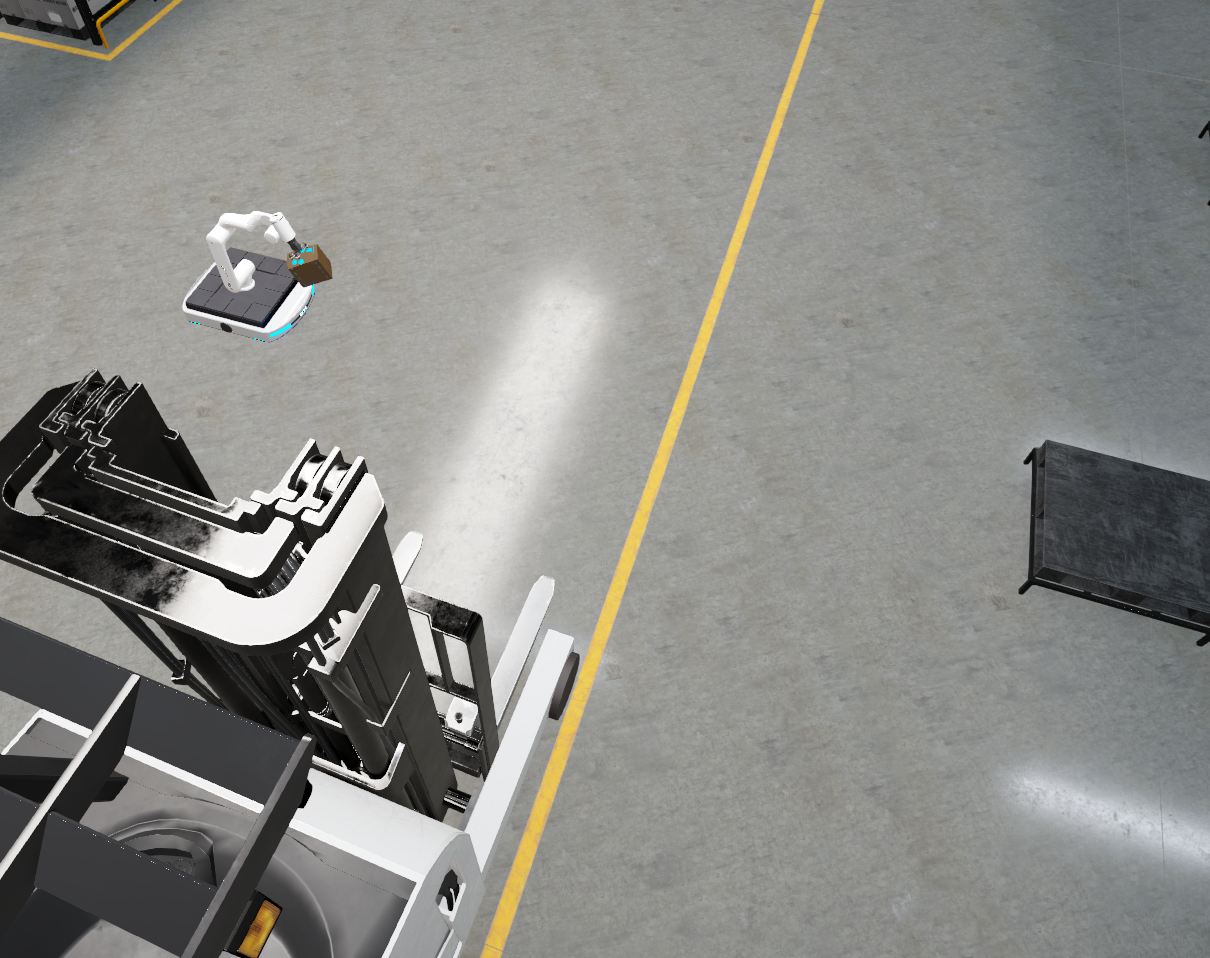}};

            \draw[red, line width=0.9mm, opacity=0.7] (-1.4, 0.) .. controls (0.2, 1.5) and (0.7, 2.) .. (0.5,-1.7);
            
            \draw[orange, ultra thick, opacity=0.7] (0.5, 0.2) -- (1.0, 1.1);
            \node[draw, orange, ultra thick, rounded corners, fill=white, opacity=0.8] at (0.2,1.4) {Greedy path};
            \draw[orange, ->, line width=0.9mm, opacity=0.7, dotted] (-1.2, 0.7) -- (1.9, -0.2);
            
            \draw[red, ->, line width=0.9mm, opacity=0.7] (-0.5, -0.4) -- (-0.15, 0.3);
            \node[draw, red, ultra thick, rounded corners, fill=white, opacity=0.8] at (1.,-1.4) {Safety margin};
            
        \end{tikzpicture}
    \end{subfigure}
    
    
    \resizebox{9cm}{!}{
    \begin{tikzpicture}
    \small
        \node (root_r0) [draw, minimum width=0.75cm, minimum height=0.75cm] {$\rightarrow$};

        \node (fb_r1) [draw, minimum width=0.75cm, minimum height=0.75cm, below=.5cm of root_r0] {?};
        \draw (root_r0.south) -- (fb_r1.north);

        \node (c_safe_r1) [draw, fill=yellow!25, shape=ellipse, left=0.6cm of fb_r1] {Safe from forklift};
        \draw (root_r0.south) -- (c_safe_r1.north);
        
        \node (b_pick_r1) [draw, minimum width=0.75cm, minimum height=0.75cm, right=1cm of fb_r1] {Select new item};
        \draw (root_r0.south) -- (b_pick_r1.north);

        \node (seq_1_r2) [draw, minimum width=0.75cm, minimum height=0.75cm, below=.5cm of fb_r1] {$\rightarrow$};
        \draw (fb_r1.south) -- (seq_1_r2.north);

        \node (seq_0_r2) [draw, minimum width=0.75cm, minimum height=0.75cm, left=2cm of seq_1_r2] {$\rightarrow$};  
        \draw (fb_r1.south) -- (seq_0_r2.north);

        \node (c_item_at_goal_r2) [draw, fill=yellow!25, shape=ellipse, left=.75cm of seq_0_r2, align=center] {Item placed};
        \draw (fb_r1.south) -- (c_item_at_goal_r2.north);

        \node (seq_2_r2) [draw, minimum width=0.75cm, minimum height=0.75cm, right=2cm of seq_1_r2] {$\rightarrow$};  
        \draw (fb_r1.south) -- (seq_2_r2.north);

        \node (b_move_r2) [draw, right=.75cm of seq_2_r2, align=center] {$\pi_1$: Move\\(to item)};
        \draw (fb_r1.south) -- (b_move_r2.north);

        \node (b_place_r3) [draw, below=.75cm of seq_0_r2, align=center] {$\pi_3$: Place\\(item)};
        \draw (seq_0_r2.south) -- (b_place_r3.north);

        \node (c_at_goal_r3) [draw, fill=yellow!25, shape=ellipse, left=.25cm of b_place_r3, align=center] {At\\goal};
        \draw (seq_0_r2.south) -- (c_at_goal_r3.north);

        \node (c_have_item_r3) [draw, fill=yellow!25, shape=ellipse, left=.25cm of c_at_goal_r3, align=center] {Have\\item};
        \draw (seq_0_r2.south) -- (c_have_item_r3.north);

        \node (c_have_item_r3_2) [draw, fill=yellow!25, shape=ellipse, right=0.8cm of b_place_r3, align=center] {Have\\item};
        \draw (seq_1_r2.south) -- (c_have_item_r3_2.north);
        
        \node (b_move_r3) [draw, right=.25cm of c_have_item_r3_2, align=center] {$\pi_1$: Move\\(to goal)};
        \draw (seq_1_r2.south) -- (b_move_r3.north);

        \node (c_near_item_r3) [draw, fill=yellow!25, shape=ellipse, below=.5cm of seq_2_r2, align=center] {Near\\item};
        \draw (seq_2_r2.south) -- (c_near_item_r3.north);

        \node (b_grasp_r3) [draw, right=.25cm of c_near_item_r3, align=center] {$\pi_2$: Grasp\\(item)};
        \draw (seq_2_r2.south) -- (b_grasp_r3.north);
    \end{tikzpicture}
    }
    \caption{\textbf{Top}: High-fidelity warehouse environment. The agent, a small mobile manipulator, must collect and deliver items while avoiding collisions with the bigger, dynamic forklift with unknown dynamics. \textbf{Bottom}: Corresponding Behavior Tree with RL controllers $\pi_1$, $\pi_2$, and $\pi_3$. Na\"ively learning the controllers with RL can result in unsafe or greedy behavior, while our method accounts for long-term BT progress constraints. Note that the lower part of the BT applies the Implicit Sequence design principle \cite{colledanchiseBehaviorTreesRobotics2017}.
    }
    \label{fig:warehouseExample}
\end{figure}

\IEEEPARstart{B}{ehavior Trees} (BTs) provide a reactive plan execution framework that switches between multiple controllers based on the current state, and are commonly used in robotics \cite{iovino2022survey} and video games \cite{islaHandlingComplexityHalo2005, biggarModularityReactiveControl2022}.
Traditionally, such trees are built by a human designer with expert knowledge of the problem, who specifies both the task-switching tree structure and the controllers in the leaf nodes of the tree ($\pi_1, \pi_2, \pi_3$ in Fig.~\ref{fig:warehouseExample}). 
Reinforcement Learning (RL), on the other hand, can learn near-optimal controllers~\cite{sutton2018reinforcement} for various domains \cite{berner2019dota, degrave2022magnetic, DBLP:journals/nature/SchrittwieserAH20, DBLP:conf/rss/KostrikovSL23} -- but suffers from a number of practical limitations: 
It is often difficult to design scalar-valued reward functions that induce the desired behavior \cite{ha2020learning, tessler2018reward, rlblogpost, DBLP:journals/aamas/VamplewSKRRRHHM22} and (safe) exploration in high-dimensional, sparse-reward tasks remains challenging \cite{pmlr-v80-riedmiller18a, chen2021decision, precup2000temporal}.

There exists mutual benefit in combining these two frameworks: From the BT perspective, RL can provide a way for learning near-optimal controllers to be used in the BT, while from the RL perspective, BTs can add structured domain knowledge and simplify the learning problem by breaking it down into smaller subproblems that can be learned separately. 
However, na\"ively learning controllers with RL as part of BTs can lead to poor performance.
Shortcomings can take the form of internal control switching oscillations \cite{kartasevImprovingPerformanceBackward2023} or suboptimal handover between controllers \cite{kartasevImprovingPerformanceLearned2023}. 
These issues stem from RL algorithms greedily optimizing rewards without considering the constraints imposed by the BT.

We suggest that these constraints are automatically identified from the BT, and then taken into account in the RL training.
For example, the BT in Fig.~\ref{fig:warehouseExample} works best if the \texttt{Place} controller $\pi_3$ remains close to the goal location, and does not prematurely drop the item, since if it moves away \texttt{Move} has to move it back, and if it drops the item in the wrong place, \texttt{Grasp} has to pick it up again.
Note that the \texttt{Move} controller $\pi_1$ appears twice in the BT, and is subject to different constraints, depending on the state. When moving towards the item, no effort is needed to avoid dropping it.

Thus, in this paper, we investigate how to integrate RL with BTs to leverage the structured domain knowledge that BTs provide while ensuring that the  RL learns controllers that respect the constraints imposed by the BT. 
Our approach introduces so-called \textit{progress constraints}, which prevent RL controllers from undoing already completed steps in the BT task, such as dropping an object that was just grasped.
We exploit theoretical convergence results for BTs to identify the constraint set~\cite{ogrenBehaviorTreesRobot2022}, and then use feasibility estimation to learn estimators for these constraints.

Thus, we present the first method for learning both RL controllers and constraints within BTs and make the following contributions:
\begin{enumerate}
    \item An extension of BT convergence concepts previously used for RL in BTs \cite{kartasevImprovingPerformanceBackward2023} using the results in \cite{ogrenBehaviorTreesRobot2022} to address a broader class of BTs. 
    \item Progress constraints: We show how to learn feasibility estimators that constrain the action set to avoid undoing the progress of BTs.
    \item We show how to use these estimators in RL algorithms, thereby providing a general-purpose learning algorithm for BTs with RL controllers.
\end{enumerate}

\section{RELATED WORK}
One of the earliest works on combining BTs and RL is \cite{pereira2015framework}, which introduces a ``learning node'' that applies na\"ive Q-learning~\cite{watkins1992q, mnih2015human}, without accounting for the implicit constraints a BT imposes on its controllers discussed in this paper.
Prior work~\cite{kartasevImprovingPerformanceBackward2023} and our results show that this approach does not respect the constraints induced by the BT, which can be attributed to the greedy, off-policy backup in Q-learning~\cite{russell2003q, laroche2017multi}.

The work that is most similar to this one is \cite{kartasevImprovingPerformanceBackward2023}.
Both papers are based on the idea of using convergence analysis of BTs to improve RL when learning controllers.
There are however two fundamental differences. First, \cite{kartasevImprovingPerformanceBackward2023} uses convergence results from \cite{ogren2020convergence} that only covers a specific class of BTs (backward chained designs) whereas this paper uses newer convergence results from \cite{ogrenBehaviorTreesRobot2022} that apply to all BTs. Second, in \cite{kartasevImprovingPerformanceBackward2023} reward shaping is used, giving the undesired transitions a large negative reward, whereas in this paper we construct progress constraints using feasibility estimators to avoid such transition all-together, independently of the reward functions. Finally, in the evaluation we compare the results to that of \cite{kartasevImprovingPerformanceBackward2023}.

Another recent work combining BT and RL \cite{kartasevImprovingPerformanceLearned2023} deals with the optimization of control switching on the boundary between two controllers.
While dealing with a similar setting as this paper, \cite{kartasevImprovingPerformanceLearned2023} addresses a different issue that aims to improve the 
handover between different controllers by giving rewards to one controller based on the preferences of the next.

A review of safe control can be found in \cite{hsu2023safety}.
We draw motivation from methods that identify safe state-space regions that controllers should not leave, i.e. keep invariant.
In particular, we make use of dynamic programming ``feasibility'' methods, as introduced in \cite{fisac2019bridging} and expanded on in \cite{yu2022reachability}, that are traditionally used  for identifying sets of safe states.
Differently from these prior works, we are applying feasibility methods not for safe control but to track and support the progress in a BT and use the resulting estimators for constraining the RL controllers.
A comprehensive review of constrained RL can be found in \cite{gu2022review}, 
but that work contains no results on how BTs can be used to identify contraints.
Finally, we apply ``action masking'' methods as described in \cite{kalweit2020deep, huang2020closer, rietz_prioritized_2023}.
However, none of those works share our BT application.

\section{BACKGROUND}
We begin with a summary of relevant background information. First, we give a formal definition of BTs in Sec.~\ref{sec:bt} and review important convergence properties of BTs in Sec.~\ref{sec:bt-convergence}. We then provide the necessary definitions from RL in Sec.~\ref{sec:RL} and lastly review feasibility analysis in Sec.~\ref{sec:feasibility}.

\subsection{Behavior Trees}
\label{sec:bt}
Let the state be $\mathbf{x} \in \mathcal{X} \subset \mathbb{R}^n$ and evolve according to the (unknown) discrete-time dynamics $\mathbf{x}_{t+1} = f(\mathbf{x}_t, \mathbf{u})$. With control actions $\mathbf{u} \in \mathcal{U} \subset \mathbb{R}^k$, the policy $\pi_i : \mathcal{X} \rightarrow \mathcal{U}$ is the controller that runs when the BT node $i$ is executing \cite{ogrenBehaviorTreesRobot2022}.

\begin{definition}[Behavior Tree]
\label{bts:thesis.thesis.def:BT}

A BT is a two-tuple 
\begin{equation}
 \bt_i=\{\pi_i,r_i\}, 
\end{equation}
where $i\in \mathbb{N}$ is the index of the tree, $\pi_i$ is the controller and 
$r_i: \mathbb{R}^n \rightarrow  \{\mathcal{R},\mathcal{S},\mathcal{F}\}$ is the return status that can output either 
\emph{Running} ($\mathcal{R}$),
\emph{Success} ($\mathcal{S}$), or
\emph{Failure} ($\mathcal{F}$).
Let the Running region ($R_i$),
Success region ($S_i$) and
Failure region ($F_i$) correspond to a partitioning of the state space:
    \begin{align*}
        R_i &= \{x: r_i(x)=\mathcal{R} \}, \\
        S_i &= \{x: r_i(x)=\mathcal{S} \}, \\
        F_i &= \{x: r_i(x)=\mathcal{F} \}. 
    \end{align*}
\end{definition}

The return status $r_i$ is used when combining BTs into larger trees. While every node in the tree can formally be considered an individual tree, we often call certain types of nodes by different names \cite{colledanchise2018behavior}. Nodes that execute controllers are called \textit{behaviors} (rectangles containing $\pi_0, \pi_1, \pi_2$ in Fig.~\ref{fig:warehouseExample}). Nodes that return only Success \textit{(true)} or Failure (\textit{false}) are known as \textit{conditions} that can be used as predicates (ellipsoids in Fig.~\ref{fig:warehouseExample}). 

For example, consider the BT in Fig.~\ref{fig:warehouseExample}. Nodes are iterated from left to right. The Sequence ($\rightarrow$) node returns \texttt{Success} if all children succeed, while the Fallback ($?$) node returns \texttt{Success} if any child succeeds. The fallback node in the figure represents a so-called implicit sequence \cite{colledanchiseBehaviorTreesRobotics2017}, with intended progression from right to left.  If the condition \texttt{Safe} (from Forklift) is unsatisfied (returns \texttt{Failure}), the whole tree returns \texttt{Failure} and an emergency stop is triggered. 
The middle subtree is executed only if the condition \texttt{Safe} is satisfied and \texttt{Item Placed} is not. 
If conditions like \texttt{At Goal} or \texttt{Have Item} change their values during execution of controller $\pi_i$, the BT automatically switches to the appropriate controller. For instance, if \texttt{Have Item} becomes unsatisfied while executing $\pi_3:$\texttt{Place} or $\pi_1:$ \texttt{Move (to goal)}, the BT will return to executing $\pi_2$ in the next time step.
 
%
%
%
%
%
%
%

\subsection{Behavior Tree convergence analysis}\label{sec:bt-convergence}
We build on the BT convergence analysis~\cite{ogren2020convergence, ogrenBehaviorTreesRobot2022} to constrain the RL controllers in a BT.
The key idea is to make sure the RL controllers keep the so-called \textit{convergence sets} invariant while striving to complete each subtask.
In order to define these sets, we first have to define so-called influence and operating regions.
The influence regions can be recursively computed considering the BT structure.
\begin{definition}(Influence region)
\label{def:influence_region}
For a given node $i$, we define $p(i)$ as the parent node of $i$ and $b(i)$ as the closest brother/sibling (having the same parent) node to the left of $i$. Based on that, the Influence region $I_i$ for node i can be determined recursively as:
    \begin{equation*}
        \begin{aligned}[c]
        I_i&=\mathcal{X}\\
        I_i&=I_{p(i)}\\
        I_i&=I_{b(i)} \cap S_{b(i)} \\
        I_i&= I_{b(i)} \cap F_{b(i)} 
        \end{aligned}
        \qquad \qquad
        \begin{aligned}[c]
        \text{If $i$ is the root}  \\
        \text{If } \nexists b(i) \And \exists p(i) \\
        \text{If $p(i)$ is a Sequence }\exists b(i)\\
        \text{If $p(i)$ is a Fallback }\exists b(i) \\
        \end{aligned}
    \end{equation*}
\end{definition}
Intuitively, the influence region $I_i$ is the part of the state-space in which subtree $i$ 
is able to influence the execution of the BT, by either its controller $\pi$ or its return status $r_i$.

Based on influence regions, we can define operating regions, which are the parts of the space where node $i$ has return status \texttt{Running} and executing its controller $\pi_i$. 
\begin{definition}(Operating Regions)
The operating region $\Omega_i$ of  node $i$, is the intersection of it's influence region $I_i$ and running regions $R_i$ 
\begin{equation}
    \Omega_i = I_i \cap R_i
\end{equation}
For a given node $i$, with operating region $\Omega_i$, the corresponding controller $\pi_i$ is executed when $\mathbf{x} \in \Omega_i$. 
The operating regions of the children of $i$ are a partitioning of $\Omega_i$.
\end{definition}

Thus, a given BT induces a partition of the state space into operating regions $\Omega_i$, where controller $\pi_i$ is executing when the state $\mathbf{x} \in \Omega_i$, for more details, see \cite{ogrenBehaviorTreesRobot2022}.

Given these concepts, we can state the main convergence result from \cite{ogrenBehaviorTreesRobot2022}, giving sufficient conditions for reaching the overall success region $S_0$, corresponding to completion of all tasks.

\begin{theorem}
\label{th_convergence}(Convergence of BTs, from \cite{ogrenBehaviorTreesRobot2022})\label{thm:convergenceBT}
Given the set of BT leaf nodes $J = \{1,...,j\}$, let 
\begin{equation} \label{eq:constraint}
C_i = ((\bigcup_{j \geq i} \Omega_j) \cup S_0).
\end{equation}    
If there exists a re-labeling of the nodes
such that for all $i \in J$,
the convergence set $C_i$ is invariant under $\pi_i$ and there exists a time-horizon $h$, such that if $\mathbf{x}_t \in \Omega_i$ then $\mathbf{x}_{t+h}\notin \Omega_i$. 
Then, the state $\mathbf{x}$ will enter the success region $S_0$ of the root of the tree (the overall success of the BT) in finite time.
\end{theorem}

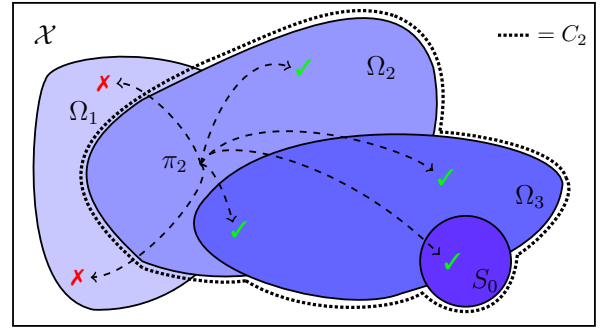
\begin{figure}[t]
    \centering
    \resizebox{0.9\linewidth}{!}{
    \begin{tikzpicture}

        \definecolor{shade1}{RGB}{200,200,255}
        \definecolor{shade2}{RGB}{150,150,255}
        \definecolor{shade3}{RGB}{100,100,255}
        \definecolor{shade4}{RGB}{100,50,255}

        \draw[thick] (-1,0) rectangle (8,5);
        \node at (-0.5,4.5) {\large $\mathcal{X}$};

        \filldraw[fill=shade1, draw=black, thick, opacity=1] 
            (-0.5,3.8) 
            .. controls (-0.4, 4.3) and (2.5,4.5).. (2.8,3) 
            .. controls (3,2) and (2, 0.2).. (0.25,0.25) 
            .. controls (-1,0.5) and (-0.7,3.5) .. cycle;
        \node at (0.1,3.3) {\large $\Omega_1$};
    
        \filldraw[fill=shade2, draw=black, thick, opacity=1] 
            (1,3.5) 
            .. controls (3,4.5) and (5,5.5) .. (5.5,4) 
            .. controls (6,1.5) and (3,0.2) .. (1,1) 
            .. controls (-0.1,2) and (-0.2,2.7) .. cycle;
        \node at (4.7,4) {\large $\Omega_2$};
    
        \filldraw[fill=shade3, draw=black, thick, opacity=1] 
            (2,2) 
            .. controls (3.2,3.5) and (7.4,3) .. (7.5,2.2) 
            .. controls (7.4,1.2) and (6,0.8) .. (5,0.5) 
            .. controls (4,0.1) and (1,1) .. cycle;
        \node at (7,2) {\large $\Omega_3$};

        \filldraw[fill=shade4, draw=black, thick, opacity=] (6,1) circle (20pt);
        \node at (6.3,0.7) {\large $S_0$};

    \node at (1.5, 2.5) {\large $\pi_2$};

    \draw[->, thick, dashed] (1.9,2.5) .. controls (2., 3) and (2.5,4.3) .. (3.3,4);
    \node at (3.5, 4) {\textcolor{green}{\ding{51}}};

    \draw[->, thick, dashed] (1.9,2.5) .. controls (2.6,3.5) and (5, 2.8) .. (5.5,2.4);
    \node at (5.7, 2.3) {\textcolor{green}{\ding{51}}};
    
    \draw[->, thick, dashed] (1.9,2.5) .. controls (2,3) and (4, 2.7) .. (5.6,1.1);
    \node at (5.8, 1) {\textcolor{green}{\ding{51}}};
    
    \draw[->, thick, dashed] (1.9,2.5) .. controls (2.1,2.5) and (2.3, 2) .. (2.4,1.6);
    \node at (2.5, 1.5) {\textcolor{green}{\ding{51}}};

    \draw[->, thick, dashed] (1.9,2.5) .. controls (1.8,3) and (1, 3.8) .. (0.6,3.75);
    \node at (0.4, 3.75) {\textcolor{red}{\ding{55}}};

    \draw[->, thick, dashed] (1.9,2.5) .. controls (2.2,2.3) and (1, 0.8) .. (0.2,0.75);
    \node at (0, 0.75) {\textcolor{red}{\ding{55}}};

    \draw[black,line width=0.5mm,densely dotted] (2,4.1) .. controls (3,4.5) and (4,5.1) .. (4.9,4.8);
    \draw[black,line width=0.5mm,densely dotted] (4.9,4.8) .. controls (5.4,4.6) and (5.8,4.1) .. (5.6,3);
    \draw[black,line width=0.5mm,densely dotted] (5.6,3) .. controls (5.9,3.) and (7.4,2.8) .. (7.55,2.3);
    \draw[black,line width=0.5mm,densely dotted] (7.55,2.3) .. controls (7.8, 1.9) and  (7, 1.2).. (6.8, 1.1);
    \draw[black,line width=0.5mm,densely dotted] (6.8, 1.1) .. controls (6.7, 0) and (5.7, 0.1) .. (5.35,0.5);
    \draw[black,line width=0.5mm,densely dotted] (5.35,0.52) .. controls (4.9, 0.22) and (3.5, 0.14) .. (2.6,0.7);
    \draw[black,line width=0.5mm,densely dotted] (2.6,0.7) .. controls (2, 0.6) and (1.1,0.85) .. (1,0.9);
    \draw[black,line width=0.5mm,densely dotted] (1,0.9) .. controls (-.4, 2.1) and (-.4, 3.) .. (2,4.1);

    \draw[black,line width=0.5mm,densely dotted] (6.5, 4.5) -- (7, 4.5);
    \node at  (7.5, 4.5) {$= C_2$};
    
    \end{tikzpicture}
    }
    \caption{Visualization of operating regions, a controller, and its convergence set. The controller $\pi_2$ may leave its operating region $\Omega_2$ but must remain in its convergence set $C_2$.}
    \label{fig:omega_regions}
\end{figure}

The intuition behind Theorem \ref{thm:convergenceBT} is as follows:
Assume some intended progression (see Remark \ref{re_intended_order}) through the operating regions $\Omega_1, \dots \Omega_J$ and let us label the leaf nodes of the BT such that the controllers for the operating regions are numbered in order of increasing intended progression. 
Then, if we can guarantee that all controllers strictly transition between operating regions according to $\Omega_j \rightsquigarrow \Omega_{j+k}, k \ge 1$, as illustrated in Fig.~\ref{fig:omega_regions}, each $C_i$ is invariant under the respective $\pi_i$ and the BT is guaranteed to reach a state $\mathbf{x} \in S_0$ in finite time.
Conversely, if any controller violates this order, it is undoing progression that has already been achieved, forcing the agent to go back and do the same task again.
The key idea of our method is make use of the 
invariance constraints from Theorem~\ref{th_convergence} on $C_i$ and imposing them on the RL controller $\pi_i$.

\begin{remark}[Intended order of progression]
\label{re_intended_order}
The intended order of progression is often clear for a manually designed BT.
If the BT was designed using the backward chaining approach \cite{colledanchiseBlendedReactivePlanning2019},
the intended order of progression can be found by a depths/left first numbering of the leaves.
Another example is the so-called implicit sequence, see \cite{colledanchise2018behavior}, and the lower part of the BT in Fig.~\ref{fig:warehouseExample}, where the intended progression would be right to left.
\end{remark}

\begin{remark}
   For the example in Figure~\ref{fig:warehouseExample} we have an intended progression from right to left due to the implicit sequence, i.e.    
   \texttt{Move(ToItem)} $\rightarrow$ 
   \texttt{Grasp} $\rightarrow$ 
   \texttt{Move(ToGoal)} $\rightarrow$ 
   \texttt{Place)}, with the corresponding convergence sets to be kept invariant  
   \begin{align*}
       C_{Move(ToItem)}=& \Omega_{Move(ToItem)} \cup  \Omega_{Grasp} \\
       &\cup  \Omega_{Move (To Goal)} \cup \Omega_{Place} \cup S_0 \\
       C_{Grasp}=&  \Omega_{Grasp} \cup  \Omega_{Move (To Goal)} \cup \Omega_{Place} \cup S_0\\
       C_{Move(To Goal)}=&  \Omega_{Move(To Goal)} \cup \Omega_{Place} \cup S_0 \\
       C_{Place}=& \Omega_{Place} \cup S_0.
   \end{align*}

\end{remark}

\subsection{Reinforcement Learning}
\label{sec:RL}
Reinforcement learning problems are formalized as Markov Decision Processes (MDPs). An MDP is a tuple $\mathcal{M} \equiv \langle \mathcal{X, U}, r, f, \gamma, \mu\rangle$, 
where $\mathcal{X}$, $\mathcal{U}$, and $f$ are aligned with the BT notation above and respectively denote the state-space, action-space, and transition function.
$r: \mathcal{X} \times \mathcal{U} \to \mathbb{R}$ is a scalar-valued reward function, $\gamma \in [0, 1]$ is a discount factor, and $\mu$ is a distribution over initial states. The goal in RL is to find a policy (henceforth referred to as \textit{controller}) $\pi: \mathcal{X} \times \mathcal{U} \to [0, 1]$ that maximizes expected, discounted return
\begin{equation}
    \mathbb{E}_{(\tau \sim \pi)}  \Big[{\sum_{t=0}^\infty \gamma^t r(\mathbf{x}_t, \mathbf{u}_t)}\Big],
    \label{eq:return_objective}
\end{equation}
where $\mathbf{x}_t \in \mathcal{X}$, $\mathbf{u}_t \in \mathcal{U}$, and $(\tau \sim \pi)$ is shorthand for denoting trajectories $\tau$ with actions sampled according to the policy and states sampled according to the transition dynamics. 

\subsection{Feasibility Estimation}
\label{sec:feasibility}
The main idea in feasibility estimation is to use dynamic programming techniques for safety analysis, as suggested in \cite{fisac2019bridging}.
Imagine we have some set of \emph{failure states} that we want to avoid because they either correspond to collisions or are undesirable in some other way. 
Then we call the complement of these states the \emph{constraint set} $\mathcal{K}\subset \mathcal{X}$.
Assuming  $\mathcal{K}$ is closed, let $l:\mathcal{X}\rightarrow \mathbb{R}$  such that $l(\mathbf{x}) > 0 \Leftrightarrow \mathbf{x}\in \mathcal{K}$ (for example, $l(\mathbf{x})$ might be a measure of the distance from $\mathbf{x}$ to the nearest failure state). 
We can then define the state feasibility function as
\begin{equation}
    V(\mathbf{x})=\sup_{\pi} \inf_{t \geq 0} l(x(t)),
\end{equation}
where $x(t)$ is a state trajectory resulting from the controller $\pi$.
Intuitively, $V(\mathbf{x})$ measures how close to the set of failure states we will come in the future, starting from $\mathbf{x}$ and applying a controller that optimally avoids the unsafe set.
It is shown in \cite{fisac2019bridging} that the state feasibility function satisfies the discrete-time Bellman equation
\begin{equation}
\label{eq_bellman1}
    V(\mathbf{x}) = \min[l(\mathbf{x}), \max_{\mathbf{u}} V( f(\mathbf{x}, \mathbf{u}) )].
\end{equation}
However, this equation does not induce a contraction, which is needed to ensure convergence. To remedy this, Fisac et al.~\cite{fisac2019bridging} introduce the following estimator 
\begin{equation}
\label{eq_bellman_gamma}
    V(\mathbf{x}) = (1-\gamma) l(\mathbf{x}) + \gamma \min[l(\mathbf{x}), \max_{\mathbf{u}} V( f(\mathbf{x}, \mathbf{u}))],
\end{equation}
which induces a contraction mapping and show that the solution converges to Eq.~(\ref{eq_bellman1}) as $\gamma \rightarrow 1$. 
Finally, the state-action feasibility function $Q$ can be found by solving
\begin{equation}
\label{eq_bellman_Q_gamma}
    Q(\mathbf{x}, \mathbf{u}) = (1-\gamma) l(\mathbf{x}) + \gamma \min[l(\mathbf{x}), \overbrace{\max_{\mathbf{u}'} Q(f(\mathbf{x}, \mathbf{u}), \mathbf{u}')}^{V(\mathbf{x'})}].
\end{equation}
Now, for any state $\mathbf{x}: V(\mathbf{x}) > 0$, there must be a non-empty set of safe actions 
\begin{equation}\label{eq:action_mask}
    \mathcal{U}(\mathbf{x}) = \{\mathbf{u}: Q(\mathbf{x}, \mathbf{u})\geq 0\},
\end{equation}
and as long as we select $\mathbf{u}_t \in \mathcal{U}(\mathbf{x})$, the next state will also be in $\mathcal{K}$.

\section{Method: Feasibility-constrained Behavior Tree Reinforcement Learning}\label{sec:method}
\subsection{Problem definition}

We now present our method for solving \textit{Progress-Constrained Behavior Tree Reinforcement Learning} (CBTRL) problems: Given a BT $\mathcal{T}_0$ with $J$ behavior nodes that partitions the state $\mathcal{X}$ into operating regions $\Omega_1, \dots, \Omega_J$, we wish to use RL to learn controllers $\pi_1, \dots, \pi_I$, $I\leq J$, such that the overall task modeled by the BT is solved successfully when the learned controllers are used in the BT's behavior nodes.
Since we often want to reuse the same controller $\pi_i$ in multiple behavior nodes, e.g. the \texttt{Move} controller in Fig.~\ref{fig:warehouseExample}, we allow $I \le J$ and denote the set of behavior nodes in which controller $i$ is used by $\mathcal{J}_i$.
For each controller $i$ that we want to learn, we define a ``subtask'' MDP $\mathcal{M}_i = \langle \mathcal{X}, \mathcal{U}, r_i, f, \gamma, \mu_i \rangle, i \in \{ 1, \dots I\}$ that shares the BTs state space $\mathcal{X}$, action space $\mathcal{U}$, and transition function $f$. 
The distribution of initial states, $\mu_i$, is a subset of the operating region $\Omega_i$ and can be induced with the controllers in $\mathcal{T}_{i-1}$, while each reward function $r_i$ can be chosen to facilitate quick learning of controller $\pi_i$ to reach the respective success region $\mathcal{S}_i$.
However, as alluded to before, directly maximizing the different $\mathcal{M}_i$ with RL does not necessarily induce convergent behavior of the overall system~\cite{kartasevImprovingPerformanceBackward2023}.

\subsection{Implementing Progress constraints}

To solve CBTRL problems, we  take the convergence analysis of BTs into account.
According to Theorem~\ref{th_convergence}, the BT is guaranteed to converge to the overall success region $S_0$ if each controller $\pi_i$ leaves its operating region $\Omega_i$ in finite time while keeping the convergence region $C_i$ invariant.
With this in mind, let $\Pi_i$ denote the set of controllers under which $C_i$ remains invariant.
While computing $\Pi_i$ directly is generally intractable, Equations~(\ref{eq_bellman_Q_gamma}, \ref{eq:action_mask}) provide a method to estimate the set of feasible actions $\mathcal{U}(\mathbf{x})$ that preserve invariance of a constraint set $\mathcal{K}$.

Therefore, for controllers used in a single behavior node, we directly set $\mathcal{K}=C_i$ and restrict the action space of $\mathcal{M}_i$ according to $\mathcal{U}_i(\mathbf{x_t})$, 
effectively projecting $\pi_i$ into $\Pi_i$.
For controllers used in multiple behavior nodes, the applicable constraint set depends on which behavior node $j \in \mathcal{J}_i$ selected the controller. 
This can be uniquely determined by identifying the active operating region $\Omega_j \subset C_i$ for the current state $\mathbf{x}$. 
Since convergence regions are nested, i.e., $C_j \subseteq C_i$ for any $j \ge i$, masking the action space according to $\mathcal{U}_j(\mathbf{x})$ still ensures that $C_i$ remains invariant.
Thus, by selecting the appropriate constraint set $C_j$ based on the active operating region and masking the action space according to $\mathcal{U}_j(\mathbf{x})$, we can use controller $\pi_i$ in multiple behavior nodes $\mathcal {J}_i$, while still guaranteeing that the corresponding convergence set $C_i$ is kept invariant.
This allows us to use any RL algorithm that supports invalid action masking \cite{kalweit2020deep, huang2020closer, rietz_prioritized_2023} to optimize $\pi_i$ while satisfying progress constraints. 
This is the core idea behind our method, based on which we derive our full learning algorithm for solving CBTRL problems.

\subsection{Learning algorithm}


Our method learns the controllers $\pi_1, \dots, \pi_I$ iteratively and separately, following the intended order of progression (Remark~\ref{re_intended_order}) in the BT $\mathcal{T}$ to ensure that subtrees $\mathcal{T}_0, \dots, \mathcal{T}_{i-1}$ and controllers are available to induce $\mu_i$ when learning controller $i$.
If there are constraints affecting even the first controller $\pi_1$, like in Fig.~\ref{fig:warehouseExample}, we assume the availability of some dataset of transitions $\mathcal{D}_0$ for training the feasibility estimator $Q_1$ for $C_1$. 
If unavailable, $\mathcal{D}_0$ can be obtained using exploration techniques like \cite{Thananjeyan2021recovery} -- more detailed data collection procedures, methods obtaining $Q_1$ analytically, or computing $Q_1$ with access to environment dynamics~\cite{ames2019control} are applicable but out of scope for this work.

Learning $Q_1$ requires a labeling function $l_1: \mathcal{X} \to \mathbb{R}$ such that $l_1(\mathbf{x})> 0 \Leftrightarrow \mathbf{x}\in \mathcal{C}_1$. 
We define $l_1$ as an binary indicator function $l_1:  1 \iff \mathbf{x} \in C_1$, which we can compute using Theorem~\ref{th_convergence} by checking whether $\mathbf{x} \in (\bigcup_{j \geq 1} \Omega_j) \cup S_0$.
We now compute $Q_1$ using the transitions in $\mathcal{D}_0$ labeled with $l_1$, following \cite{fisac2019bridging} by performing SGD w.r.t. the parameter $\phi_1$ on the Bellman feasibility residual from Equation (\ref{eq_bellman_Q_gamma}).
More generally, for some constraint $j$, this means minimizing the loss
\begin{equation}
\begin{split}
  L(\phi_j, \mathcal{D}) &= \mathbb{E}_{(\mathbf{x, u, x'})\sim\mathcal{D}}\Big[\big(\bar{y} - Q_j(\mathbf{x}, \mathbf{u}, \phi_j)\big)^2\Big], 
  \\
  \bar{y} = (1 - \gamma) & l_j(\mathbf{s}) + \gamma \min \Big(l_j(\mathbf{x}), \underset{\mathbf{u}'}{\max}\ Q_j(\mathbf{x'}, \mathbf{u}', \phi_j) \Big),   
\end{split}
\label{eq:feasibility_dqn_loss}
\end{equation}
With access to $Q_1$, we can learn $\pi_1$ by optimizing $\mathcal{M}_1$ while masking the action space according to $\mathcal{U}_1$ and storing the collected transitions as $\mathcal{D}_1$.

We follow the same scheme in all subsequent iterations of our method. 
In iteration $j$, we first train the feasibility estimator $Q_j$ using all data available thus far, i.e. $\mathcal{D}_j = \mathcal{D}_0 \cup \dots \cup \mathcal{D}_{j-1}$, then train controller $\pi_j$ using the action mask $\mathcal{U}_j$ induced from $Q_j$. 
If controller $\pi_j$ is shared with some previous behavior nodes $\mathcal{J}_j =\{l,\dots,k \} \le j$, the action space is masked based on the operating region $\Omega_k$ for each state.

Since the $i$-th feasibility estimator $Q_i$ and RL policy $\pi_i$ can be learned independently from prior components, the overall complexity of this approach grows only linearly with $I$. Lower priority RL controllers with higher indices are subject to increasingly many constraints, which naturally restricts their freedom over actions. This is by design of the priority induced by the BT and does not increase the computational complexity or difficulty of the RL objective, in fact, the smaller action space simplifies exploration.

To summarize, for each RL controller in $\{ \pi_1, \dots \pi_I\}$, we first produce the feasibility estimator for the corresponding convergence region $C_i$ and then learn the RL controller $\pi_i$ subject to that constraint. 

\begin{remark}[Customization]
Since each controller $\pi_i$ in a BT operates within a fixed region $\Omega_i$, we are not restricted to using learning for all controllers. There can be a mix or RL and model based controllers. The constraint estimator $Q_j$ can be trained independently of how the controllers in the rest of the tree function. The constraints prevent transitions into the local failure region of each controller, making their learning modular. Furthermore, due to the off-policy nature of the feasibility estimation, if prior data exists or the constraint can be inferred from a model, it can be applied directly without using the iterative process described above.
\end{remark}

\section{EXPERIMENTS}
\subsection{Experimental setup}

\noindent\textbf{Baselines}:
\begin{itemize}
    \item \textit{Standard RL}: A standard RL agent that receives -1 reward for each subtask that is \textit{not} achieved at step $t$. Since this baseline features no BT, it reveals how the structured domain knowledge from the BT can affect the final agent.
    \item \textit{BTRL (DQN / PPO)}: Proposed by \cite{pereira2015framework} as a Learning Action Node. This method employs standard, unconstrained RL principles to learn the controllers in the BT, using the environment subtask rewards, and can be seen as an ablation of our method that does not use progress constraints.
    \item \textit{BT-Penalty}: A reward penalty is added to transitions that violate progress constraints when learning the controllers in the BT with RL. This is similar to the method used in \cite{kartasevImprovingPerformanceBackward2023} but without presuming a positive reward structure for the task reward.
\end{itemize}

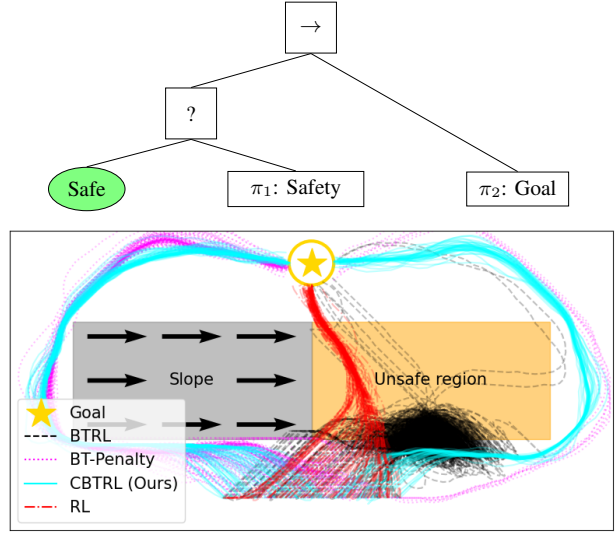
\begin{figure}
    \centering
     \resizebox{0.95\linewidth}{!}{
    \begin{tikzpicture}[scale=0.83, every node/.style={font=\small}]
        \node (root) [draw, minimum width=0.75cm, minimum height=0.75cm] {$\rightarrow$};
        \node (fallback_safe) [draw, below left=0.5cm and 1cm of root, minimum width=0.75cm, minimum height=0.75cm] {?};
        \node (safe_post) [draw, fill=green!50, shape=ellipse, below left=0.5cm and 0.75cm of fallback_safe] {Safe};
        \node (safe_policy) [draw, right=1.5cm of safe_post, minimum width=2cm, minimum height=0.5cm] {$\pi_1$: Safety};
        \node (goal_policy) [draw, right=1.5cm of safe_policy, minimum width=1.5cm, minimum height=0.5cm] {$\pi_2$: Goal};

        \draw (root.south) -- (fallback_safe.north);
        \draw (root.south) -- (goal_policy.north);
        \draw (fallback_safe.south) -- (safe_post.north);
        \draw (fallback_safe.south) -- (safe_policy.north);
        
        \node[inner sep=0pt, below=2.5cm of root] (img)
        {\includegraphics[width=.5\textwidth]{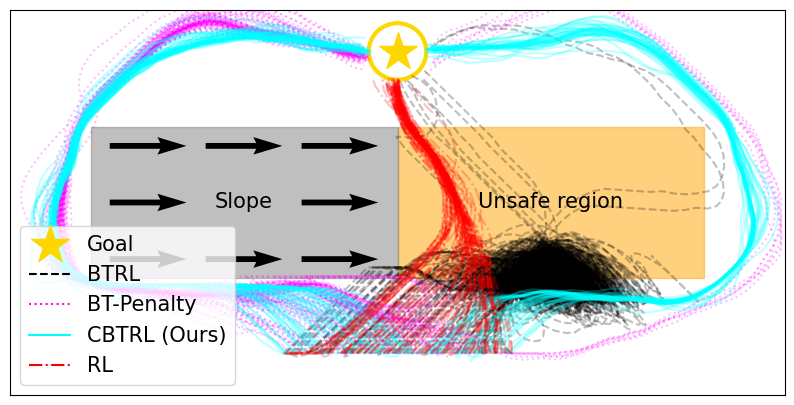}};
    \end{tikzpicture}
    }
    
    \caption{2D goal-reach environment. \textbf{Top}: Task-switching BT with RL-based controllers. The BT selects the \texttt{goal} controller $\pi_2$, when the state is in its convergence set $C_2$, which corresponds to states outside of the unsafe region and the slope.
    \textbf{Bottom}: Trajectories generated by different methods. 
    Our method converges to a near-optimal behavior for the BT task of reaching the goal without entering the unsafe region. Note that with our method, $\pi_2$ never violates the progress constraint ``safe'' during learning.
    }
    \label{fig:bt_compare_numpy}
\end{figure}

\noindent\textbf{2D goal-reach environment}: 

This environment, shown in Fig.~\ref{fig:bt_compare_numpy}, features a point-mass agent that has to navigate to a fixed goal location while avoiding an unsafe region. The state-space is  $\mathbb{R}^4$ and contains the agent's $x$ and $y$ position and velocity. The discrete action space is of size 25 and corresponds to accelerations at differing angles and magnitudes. Episodes are truncated after 200 steps. When the agent is on the slope, actions have no effect, simulating the agent sliding into the unsafe region.
The slope serves to illustrate the need for temporally extended feasibility estimation and progress constraints, since stepping onto the slope results in an inevitable but delayed constraint violation. 
While knowledge about the unsafe region is encoded in the BT, due to the model-free setting, the agent must learn about the dynamics of the slope by exploration.
The subtask reward functions for which we learn controllers are as follows:
\begin{itemize}
    \item $r_1$ Safe: -1 reward for each step spent in the unsafe area, otherwise 0.
    \item $r_2$ Goal: A negative term inversely proportional to the distance from the goal.
\end{itemize}
\textbf{Warehouse environment}:

To test the method's scalability to more complex environments, we implemented the example seen in Fig.~\ref{fig:warehouseExample}. 
Here, the agent is a mobile manipulator that has to pick up and deliver items while avoiding collision with a dynamic forklift, for which the dynamics are unknown.
The environment features an observation space of 37 continuous vector features containing the position of the manipulator's joints in the body frame, the state of the suction cup, the pose and velocity of the agent and forklift, as well as the varying positions of the box and goal locations in the body frame. The action space of 6250 discrete actions represents a product space of the possible controls for the wheels, arms, and suction cup of the agent. The subtask policies \texttt{Move}, \texttt{Grasp}, and \texttt{Place} have the following reward functions.

\begin{itemize}
    \item $r_1$ Move: A positive term proportional to the decrease in the agent's distance to the goal and a negative term for colliding with the static environment (walls, shelves).
    \item $r_2$ Grasp: A positive term proportional to the decrease of the distance of the arm to the item and a second term proportional to the decrease of the item's distance to the desired grasping position. 
    \item $r_3$ Place: A positive term proportional to decreasing the box's distance to the placing position.  
\end{itemize}
The MDPs also use a small negative penalty for passing time. 

This environment only uses the BTRL, BT-Penalty, and CBTRL methods, as the Standard RL policy did not converge to a solution within the project's computational constraints. To demonstrate the effect of the constraint, we also test applying the constraint to the BTRL and BT-Penalty methods without training them explicitly with the constraint active, as is done in CBTRL. All methods use PPO to train the policy and store the experiences in a dataset for later use during the training of the feasibility estimators. Each controller is trained separately in a dedicated learning environment to reduce the complexity of the learning process. There is also a dynamic obstacle in the form of the Forklift, which makes the ``Safe'' condition depend on the changing forklift position.

For evaluation, we created a scenario that requires a combination of all the separately trained controllers in a BT using task settings analogous to the training tasks in a previously unforeseen configuration. The results of this evaluation can be seen in Table~\ref{tab:warhouseEval}.

\subsection{Results and discussion}\label{sec:results}

We now analyze our proposed method, CBTRL in comparison to the above-mentioned baselines on our two environments. 
We focus on comparisons with respect to success rate, progress constraint violation, and sample complexity.

\begin{figure}[t]
    \centering
    \includegraphics[width=0.97\linewidth]{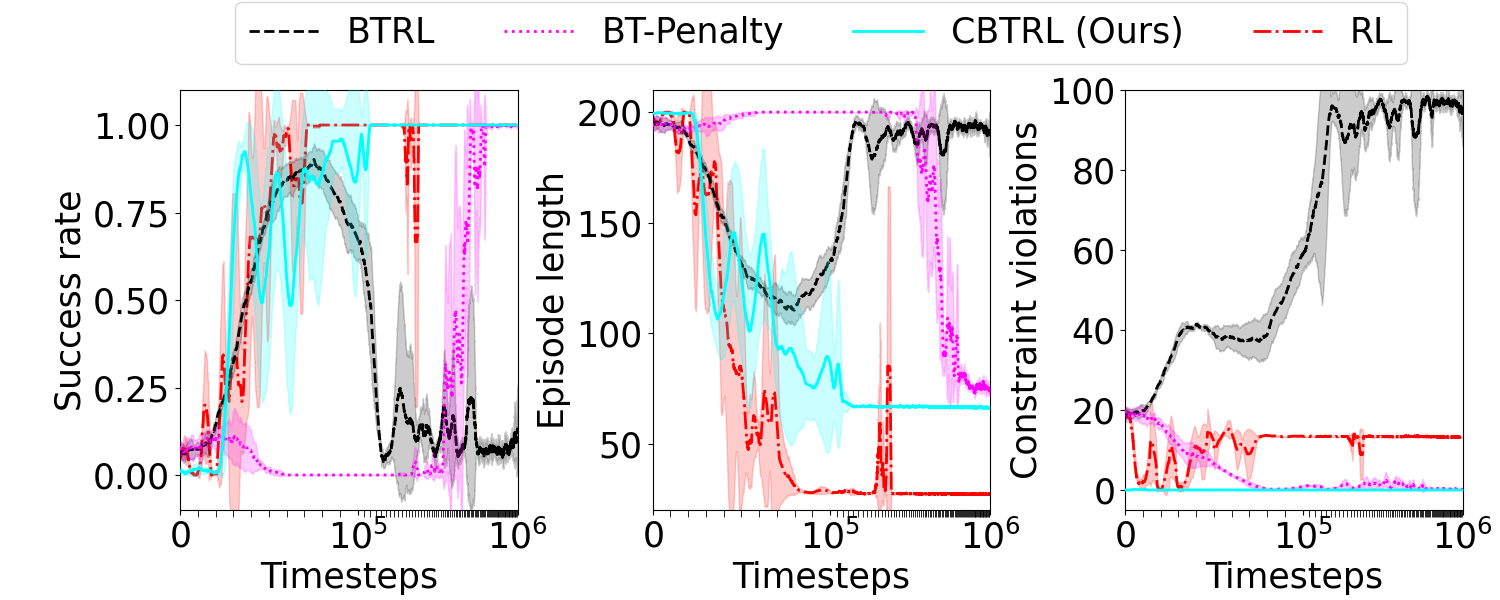}
    \caption{Empirical results on 2D navigation environment. We plot the mean of five repetitions with varying random seeds, the shaded area corresponds to one standard deviation around the mean. To enhance readability, the x-axis is linearly scaled in the range $[0, 10^5]$ and logarithmically scaled from $10^5$ to $10^6$.}
    \label{fig:2d-numpy-metrics}
\end{figure}
\begin{figure}
    \centering
    \includegraphics[width=0.95\linewidth]{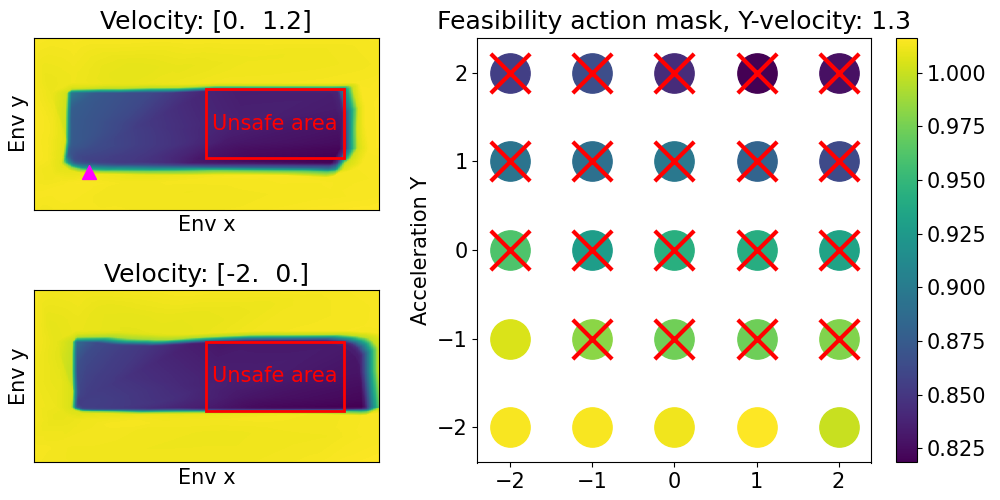}
    \caption{\textbf{Left}: Learned state feasibility function for the unsafe region in the 2D environment accounts for both agent and environment dynamics. \textbf{Right}: Induced action space mask $\mathcal{U}_1$ with the agent placed at \includegraphics[height=7pt]{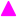} and velocities as in the top image on the left. Even though the \texttt{safety} constraint violation only occurs in the unsafe area, the constraint prevents the agent from stepping onto the slope.}
    \label{fig:feasibility-estimator}
\end{figure}

\paragraph{\textbf{Progress constraints induce RL controllers that successfully solve the BT tasks}}
\begin{figure}[t]
    \centering
    \begin{subfigure}[t]{0.93\columnwidth}
        \includegraphics[width=\linewidth]{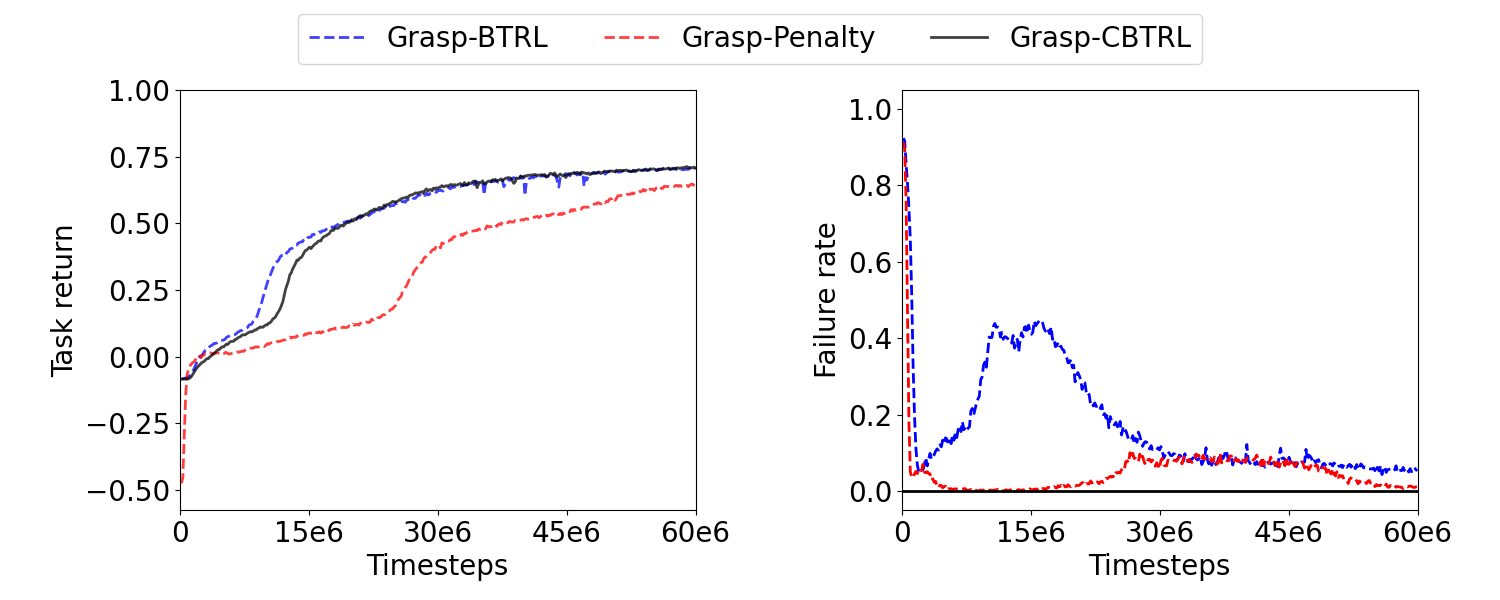}
    \end{subfigure}
        \begin{subfigure}[t]{0.93\columnwidth}
        \includegraphics[width=\linewidth]{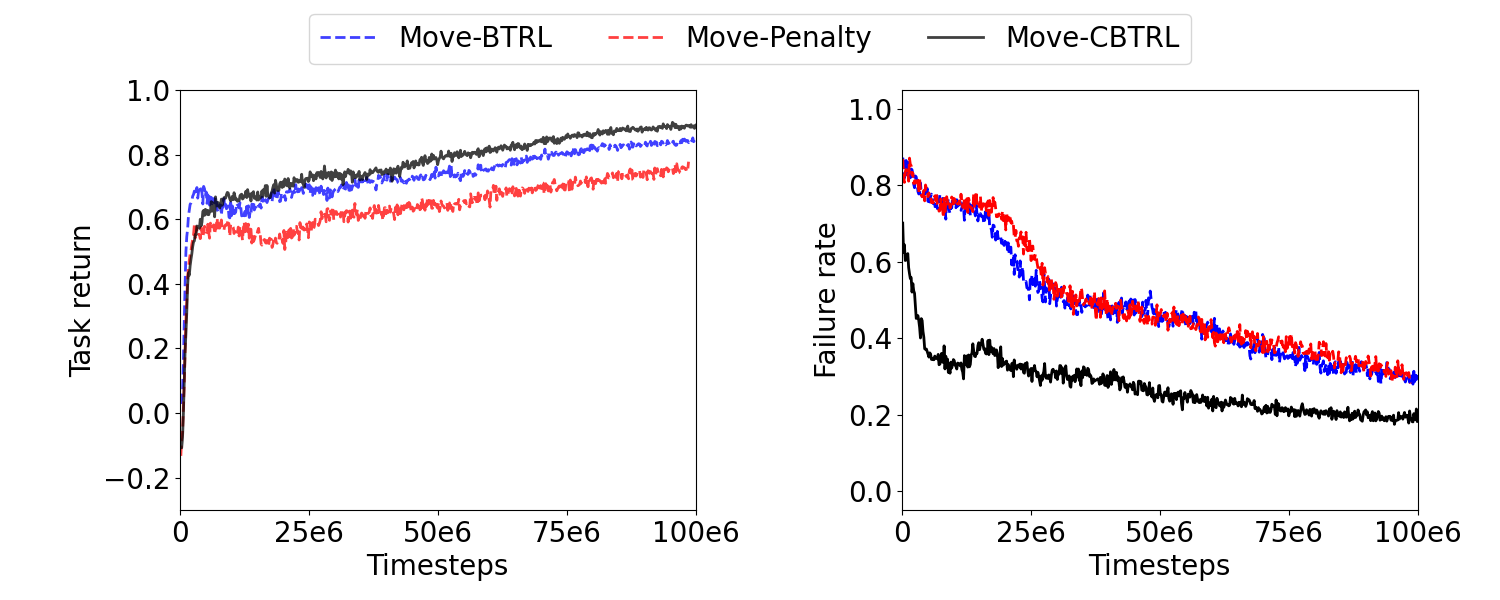}
    \end{subfigure}
        \begin{subfigure}[t]{0.93\columnwidth}
        \includegraphics[width=\linewidth]{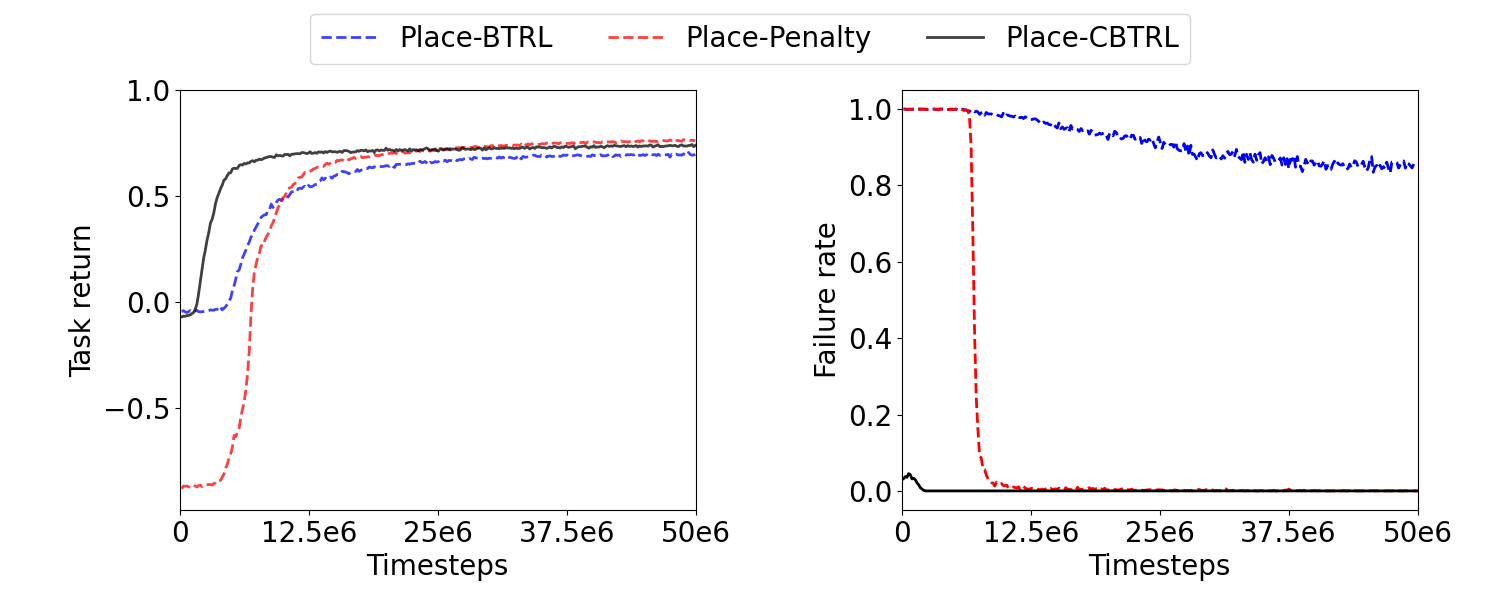}
    \end{subfigure}
    \caption{Training results of the controllers in the Warehouse environment (\texttt{Grasp} (top), \texttt{Move} (middle), \texttt{Place} (bottom)).}
    \label{fig:warehouseMetrics}
\end{figure}

\begin{table*}[t]
    \centering
    \resizebox{0.95\linewidth}{!}{
    \begin{tabular}{l|c|c|c|c|c|c|c|c|c|}
       Evaluated& Steps to &    Success &  Timeout & Failure  & Have Item & Near Item &  At Goal & Safe \\
       Method   & Complete &   Rate     & Rate     & Rate     & Violations \# : \%  & Violations \# : \% & Violations \# : \% &Violations \# : \% \\
       \hline
       BTRL     & $1186.99 \pm 959.33$ & 78\% & 3\% & 19\%   & 1211 : 54.55\% & 0 : 0\% & 62 : 2.99\% & 400 : 37.66\% \\
       BTRL \textbf{w. pc} & $2782.40 \pm 1842.94$ & 44\% & 31\% & 25\%   & 2 : 0.20\% & 0 : 0\% & 99 : 1.30\% & 389 : 34.37\%  \\
       BT-Penalty      & $1154.91 \pm 624.51$ & 83\% & 1\% & 16\%   & 155 : 13.77\% & 0 : 0\% & 1 : 0.10\% & 481 : 43.46\% \\
       BT-Penalty \textbf{w. pc}  & $1208.41 \pm 500.16$ & 82\% &1\% & 17\% & 11 : 1.10\% & 0 : 0\% & 1 : 0.10\% & 296 : 27.27\%\\
       CBTRL (ours)   & $\mathbf{1005.59 \pm 302.97}$ & $\mathbf{93\%}$ & $\mathbf{0\%}$ & $\mathbf{7\%}$ & 0 : $\mathbf{0\%}$ & 0 : 0\% & 0 : $\mathbf{0\%}$ & 198 : \textbf{19.30\%}\\
    \end{tabular}
    }
    \caption{Agent evaluation in the warehouse environment. Values are given over 1000 random episodes on an unseen but analogous setting. $\textbf{w. pc}$ indicates evaluation \textit{\textbf{w}ith \textbf{p}rogress \textbf{c}onstraint}, where the constraint is applied to the resulting controllers after training but not during training. Violations given as total number (\#) and percentage of episodes that had violations (\%)}
    \label{tab:warhouseEval}
\end{table*}

It can be seen that progress constraints have a strong effect on the agent, both during learning and in the final behavior.
For the 2D environment, Fig.~\ref{fig:bt_compare_numpy} shows trajectories of the final agents, while Fig.~\ref{fig:2d-numpy-metrics} shows metrics collected during the training of $\pi_2$.
CBTRL solves the task modeled by the BT much faster than the baselines and, crucially, without violating constraints even during training.
Standard RL, without the task-switching BT structure, learns to drive the shorter path through the unsafe region.
While the penalty-based method also learns to solve the BT task, it initially incurs constraint violations and learns slower, as the large penalties have a destabilizing effect on the RL training.
Lastly, the unconstrained BTRL method~\cite{pereira2015framework} does not account for progress constraints and gets stuck in a task-switching loop: Underneath the unsafe region, $\pi_2$ attempts to drive upwards towards the goal and enters the unsafe region. Inside the unsafe region, the $\pi_1$ safety controller is selected, driving down, out of the unsafe region. Now $\pi_2$ again attempts to drive upwards to the goal, and the cycle repeats.
This is a typical problem that might occur when doing task-switching.

A similar problem exists in the Warehouse scenario results of Table~\ref{tab:warhouseEval}.
The individually trained BTRL-Move behavior causes \textit{Have Item} violations by dropping the box. 
This causes the system to chatter between the controllers, making little progress. 
This behavior is reduced in BT-Penalty, which learns to prevent the negative reward from the \textit{Have Item} violation, and is eliminated in the CBTRL method, which prohibits releasing the box.
Applying the constraints after training to BTRL and BT-Penalty also eliminates such violations (rows 2 and 4), but induces problems discussed in the next paragraph.
%

\paragraph{\textbf{Ad-hoc constraints degrade performance and introduce failure modes}}
Learning RL controllers decoupled from progress constraint estimators and enforcing the constraints only at inference time might make sense from a modularity perspective, but leads to suboptimal performance and inconsistent behaviors. 
Table~\ref{tab:warhouseEval} shows that applying progress constraints to controllers learned in unconstrained fashion (BTRL w.pc and BT-Penalty w.pc) can drastically lower respective constraint violations. 
However, this comes at the cost of considerably increased timeout rates and reduced overall task completion. This problem stems from the fact that the policy was never trained under the constraint-aware action space. 
For example, the BTRL-Place controller might learn to complete its subtask by dropping the item from an unsafe height -- a behavior that becomes invalid once the progress constraint is enforced. 
Since the controller never experienced this restriction during training, it fails by exceeding the mission time limit when its preferred behavior is disallowed during inference. 
The CBTRL method, in contrast, integrates constraints during training, thereby forcing controllers to learn feasible strategies from the outset, leading to consistent and successful task execution. 
This demonstrates that progress constraints must be incorporated throughout training, not merely applied afterward, in order to achieve the desired behavior within BTs.

\paragraph{\textbf{Feasibility estimation captures long-term consequences of actions w.r.t progress constraints}}
The importance of long-term feasibility estimation w.r.t BT progress constraints can be seen in Fig.~\ref{fig:feasibility-estimator}. 
While the safety constraint violation only occurs when the agent enters the unsafe region,
The use of the Bellman feasibility equations also identifies the ``slope'' area as infeasible, i.e. as outside of the convergence set $C_2$ for $\pi_2$.
This is correct since once the agent steps onto the slope, it will inevitably slide into the unsafe area. 
If we were to mask the action space by forbidding only those actions that immediately violate the progress constraints, stepping onto the slope would be wrongfully allowed.
Thus, the visualization in Fig.~\ref{fig:feasibility-estimator} shows that feasibility estimation can be used to capture the long-term dynamics w.r.t progress constraints, which is needed to estimate the convergence region $C_i$ for the controllers in the BT.

\paragraph{\textbf{BT task decomposition with progress constraints reduces RL sample complexity}}
The structured domain knowledge in a BT seems to simplify the RL learning problem. 
Beyond the general benefits associated with task decomposition, the method further seems to speed up learning as the progress constraints remove parts of the action space.
This limits the policy search space since we prevent the subtask controllers $\pi_i$ from exploring solutions we already know to be suboptimal.

The effect can be most prominently seen in Fig.~\ref{fig:warehouseMetrics} during the learning of the Place behaviors in the Warehouse environment.
We see that the constrained \textit{CBTRL-Place} behavior, which is constrained by the estimator trained with data from the \textit{BTRL-Grasp} behavior,  converges to a stable result after around $13 \times 10^6$ experiences; much faster than \textit{BTRL-Place}, which needs about $22 \times 10^6$ experiences. 
This is because all actions in the product space where the box is released are removed and do not have to be explored. 
This can be verified by the failure rate of zero during the training of "CBTRL-Place".
Similarly, in Fig.~\ref{fig:2d-numpy-metrics}, CBTRL converges after roughly $10^5$ steps, while the BT-penalty method requires roughly $10^6$ steps.

\paragraph{\textbf{Feasibility estimator overhead}}
For some BTs, like the one in Fig.~\ref{fig:warehouseExample}, where the first policy, $\pi_1$: Move, is subject to a constraint, ``Safe from forklift'', training the feasibility estimator requires access to an a priori dataset.
We used 5M transitions collected by the BTRL baseline for training the first estimator for CBTRL-Move. 
Compared to the 100M transitions for the policy itself, the amount of additional data required is only 5\% extra of what is required for the policy.
For the BT in Fig.~\ref{fig:bt_compare_numpy}, no additional data was needed, since the feasibility estimator for $\pi_2$: Goal can be trained using the transition collected during learning of $\pi_1$: Safety.
The overhead from the feasibility estimators varies by domain and BT, but is generally small compared to RL's high data demands. A detailed analysis of data requirements for feasibility estimation is interesting, but beyond this work’s scope.

\section{CONCLUSIONS}
The main contribution of this paper is an algorithm for learning controllers in a BT with RL.
We propose novel \textit{progress} constraints based on feasibility estimation and BT convergence properties, which are then used to learn the controllers in the BT with constrained RL.
A limitation of this work lies in potentially insufficient data collection for learning the progress constraints, since we assume that the data collected during exploration with RL suffices for training robust feasibility estimators. 
We see a more informed and optimized way of gathering data for learning these estimators, or the integration of model-based progress constraints, as important future work.



\printbibliography

@inproceedings{rietz_prioritized_2023,
  title={Prioritized Soft Q-Decomposition for Lexicographic Reinforcement Learning},
  author={Rietz, Finn and Heinrich, Stefan and Schaffernicht, Erik and Stork, Johannes A},
  booktitle={The Twelfth International Conference on Learning Representations (ICLR2023)},
  year={2024}
}

@article{berner2019dota,
  title={Dota 2 with large scale deep reinforcement learning},
  author={Berner, Christopher and Brockman, Greg and Chan, Brooke and Cheung, Vicki and D{\k{e}}biak, Przemys{\l}aw and Dennison, Christy and Farhi, David and Fischer, Quirin and Hashme, Shariq and Hesse, Chris and others},
  journal={arXiv preprint arXiv:1912.06680},
  year={2019}
}

@article{biggarModularityReactiveControl2022,
  title = {On Modularity in Reactive Control Architectures, with an Application to Formal Verification},
  author = {Biggar, Oliver and Zamani, Mohammad and Shames, Iman},
  year = {2022},
  month = may,
  journal = {ACM Transactions on Cyber-Physical Systems (TCPS)},
  urldate = {2021-05-18}
}

@phdthesis{colledanchiseBehaviorTreesRobotics2017,
  title = {Behavior {{Trees}} in {{Robotics}}},
  author = {Colledanchise, Michele},
  year = {2017},
  urldate = {2018-11-02},
  langid = {english},
  school = {KTH Royal Institute of Technology}
}

@inproceedings{colledanchiseBlendedReactivePlanning2019,
  title = {Towards {{Blended Reactive Planning}} and {{Acting}} Using {{Behavior Trees}}},
  booktitle = {{{IEEE Int}}. {{Conference}} on {{Robotics}} and {{Automation}}},
  author = {Colledanchise, Michele and Almeida, Diogo and {\"O}gren, Petter},
  year = {2019},
  month = may,
  publisher = {IEEE},
  address = {Montreal, Canada},
  urldate = {2018-10-31}
}

@inproceedings{islaHandlingComplexityHalo2005,
  title = {Handling {{Complexity}} in the {{Halo}} 2 {{AI}}},
  booktitle = {Proceedings of the {{Game Developers Conference}} ({{GDC}})},
  author = {Isla, D.},
  year = {2005},
  urldate = {2019-06-06}
}

@inproceedings{kartasevImprovingPerformanceBackward2023,
  title = {Improving the {{Performance}} of {{Backward Chained Behavior Trees}} That Use {{Reinforcement Learning}}},
  booktitle = {{{IEEE}} {{IROS}}},
  author = {Karta{\v s}ev, Mart and Saler, Justin and {\"O}gren, Petter},
  year = {2023},
  eprint = {2112.13744},
  primaryclass = {cs, eess},
  urldate = {2023-02-12},
  archiveprefix = {arxiv},
  langid = {english},

}

@ARTICLE{kartasevImprovingPerformanceLearned2023,
  author={Kartašev, Mart and Ögren, Petter},
  journal={IEEE Robotics and Automation Letters}, 
  title={Improving the Performance of Learned Controllers in Behavior Trees Using Value Function Estimates at Switching Boundaries}, 
  year={2024},
  
  doi={10.1109/LRA.2024.3382477}}

@article{ogrenBehaviorTreesRobot2022,
  title = {Behavior {{Trees}} in {{Robot Control Systems}}},
  author = {{\"O}gren, Petter and Sprague, Christopher I.},
  year = {2022},
  journal = {Annual Review of Control, Robotics, and Autonomous Systems},
  volume = {5},
  number = {1},
  doi = {10.1146/annurev-control-042920-095314},
  urldate = {2022-03-22}
}

@article{iovino2022survey,
  title={A survey of behavior trees in robotics and ai},
  author={Iovino, Matteo and Scukins, Edvards and Styrud, Jonathan and {\"O}gren, Petter and Smith, Christian},
  journal={RAS},
  volume={154},
  year={2022},
  publisher={Elsevier}
}

@book{colledanchise2018behavior,
  title={Behavior trees in robotics and AI: An introduction},
  author={Colledanchise, Michele and {\"O}gren, Petter},
  year={2018},
  publisher={CRC Press}
}

@book{sutton2018reinforcement,
  title={Reinforcement learning: An introduction},
  author={Sutton, Richard S and Barto, Andrew G},
  year={2018},
  publisher={MIT press}
}

@article{ogren2020convergence,
  title={Convergence analysis of hybrid control systems in the form of backward chained behavior trees},
  author={{\"O}gren, Petter},
  journal={IEEE Robotics and Automation Letters},
  volume={5},
  number={4},
  pages={6073--6080},
  year={2020},
  publisher={IEEE}
}

@article{pereira2015framework,
  title={A framework for constrained and adaptive behavior-based agents},
  author={Pereira, Renato de Pontes and Engel, Paulo Martins},
  journal={arXiv preprint arXiv:1506.02312},
  year={2015}
}

@misc{baselines,
  author = {Dhariwal, Prafulla and Hesse, Christopher and Klimov, Oleg and Nichol, Alex and Plappert, Matthias and Radford, Alec and Schulman, John and Sidor, Szymon and Wu, Yuhuai and Zhokhov, Peter},
  title = {OpenAI Baselines},
  year = {2017},
  publisher = {GitHub},
  journal = {GitHub repository},
  howpublished = {\url{https://github.com/openai/baselines}},
}

@article{watkins1992q,
  title={Q-learning},
  author={Watkins, Christopher JCH and Dayan, Peter},
  journal={Machine learning},
  volume={8},
  pages={279--292},
  year={1992},
  publisher={Springer}
}

@inproceedings{russell2003q,
  title={Q-decomposition for reinforcement learning agents},
  author={Russell, Stuart J and Zimdars, Andrew},
  booktitle={Proceedings of the 20th international conference on machine learning (ICML-03)},
  pages={656--663},
  year={2003}
}

@inproceedings{yu2022reachability,
  title={Reachability constrained reinforcement learning},
  author={Yu, Dongjie and Ma, Haitong and Li, Shengbo and Chen, Jianyu},
  booktitle={International Conference on Machine Learning},
  pages={25636--25655},
  year={2022},
  organization={PMLR}
}

@inproceedings{fisac2019bridging,
  title={Bridging hamilton-jacobi safety analysis and reinforcement learning},
  author={Fisac, Jaime F and Lugovoy, Neil F and Rubies-Royo, Vicen{\c{c}} and Ghosh, Shromona and Tomlin, Claire J},
  booktitle={2019 International Conference on Robotics and Automation (ICRA)},
  pages={8550--8556},
  year={2019},
  organization={IEEE}
}

@article{kalweit2020deep,
  title={Deep constrained q-learning},
  author={Kalweit, Gabriel and Huegle, Maria and Werling, Moritz and Boedecker, Joschka},
  journal={arXiv preprint arXiv:2003.09398},
  year={2020}
}

@article{mnih2015human,
  title={Human-level control through deep reinforcement learning},
  author={Mnih, Volodymyr and Kavukcuoglu, Koray and Silver, David and Rusu, Andrei A and Veness, Joel and Bellemare, Marc G and Graves, Alex and Riedmiller, Martin and Fidjeland, Andreas K and Ostrovski, Georg and others},
  journal={nature},
  volume={518},
  number={7540},
  pages={529--533},
  year={2015},
  publisher={Nature Publishing Group UK London}
}

@inproceedings{ames2019control,
  title={Control barrier functions: Theory and applications},
  author={Ames, Aaron D and Coogan, Samuel and Egerstedt, Magnus and Notomista, Gennaro and Sreenath, Koushil and Tabuada, Paulo},
  booktitle={2019 18th European control conference (ECC)},
  pages={3420--3431},
  year={2019},
  organization={IEEE}
}

@article{ha2020learning,
  title={Learning to walk in the real world with minimal human effort},
  author={Ha, Sehoon and Xu, Peng and Tan, Zhenyu and Levine, Sergey and Tan, Jie},
  journal={arXiv preprint arXiv:2002.08550},
  year={2020}
}

@article{tessler2018reward,
  title={Reward constrained policy optimization},
  author={Tessler, Chen and Mankowitz, Daniel J and Mannor, Shie},
  journal={arXiv preprint arXiv:1805.11074},
  year={2018}
}

@misc{rlblogpost,
    title={Deep Reinforcement Learning Doesn't Work Yet},
    author={Irpan, Alex},
    howpublished={\url{https://www.alexirpan.com/2018/02/14/rl-hard.html}},
    year={2018}
}

@article{degrave2022magnetic,
  title={Magnetic control of tokamak plasmas through deep reinforcement learning},
  author={Degrave, Jonas and Felici, Federico and Buchli, Jonas and Neunert, Michael and Tracey, Brendan and Carpanese, Francesco and Ewalds, Timo and Hafner, Roland and Abdolmaleki, Abbas and de Las Casas, Diego and others},
  journal={Nature},
  volume={602},
  number={7897},
  pages={414--419},
  year={2022},
  publisher={Nature Publishing Group}
}

@article{DBLP:journals/nature/SchrittwieserAH20,
  author       = {Julian Schrittwieser and
                  Ioannis Antonoglou and
                  Thomas Hubert and
                  Karen Simonyan and
                  Laurent Sifre and
                  Simon Schmitt and
                  Arthur Guez and
                  Edward Lockhart and
                  Demis Hassabis and
                  Thore Graepel and
                  Timothy P. Lillicrap and
                  David Silver},
  title        = {Mastering Atari, Go, chess and shogi by planning with a learned model},
  journal      = {Nat.},
  volume       = {588},
  number       = {7839},
  pages        = {604--609},
  year         = {2020},
  url          = {https://doi.org/10.1038/s41586-020-03051-4},
  doi          = {10.1038/S41586-020-03051-4},
  bibsource    = {dblp computer science bibliography, https://dblp.org}
}

@inproceedings{DBLP:conf/rss/KostrikovSL23,
  author       = {Ilya Kostrikov and
                  Laura M. Smith and
                  Sergey Levine},
  editor       = {Kostas E. Bekris and
                  Kris Hauser and
                  Sylvia L. Herbert and
                  Jingjin Yu},
  title        = {Demonstrating {A} Walk in the Park: Learning to Walk in 20 Minutes
                  With Model-Free Reinforcement Learning},
  booktitle    = {Robotics: Science and Systems XIX, Daegu, Republic of Korea, July
                  10-14, 2023},
  year         = {2023},
  url          = {https://doi.org/10.15607/RSS.2023.XIX.056},
  doi          = {10.15607/RSS.2023.XIX.056},
  bibsource    = {dblp computer science bibliography, https://dblp.org}
}

@article{DBLP:journals/aamas/VamplewSKRRRHHM22,
  author       = {Peter Vamplew and
                  Benjamin J. Smith and
                  Johan K{\"{a}}llstr{\"{o}}m and
                  Gabriel de Oliveira Ramos and
                  Roxana Radulescu and
                  Diederik M. Roijers and
                  Conor F. Hayes and
                  Fredrik Heintz and
                  Patrick Mannion and
                  Pieter J. K. Libin and
                  Richard Dazeley and
                  Cameron Foale},
  title        = {Scalar reward is not enough: a response to Silver, Singh, Precup and
                  Sutton {(2021)}},
  journal      = {Auton. Agents Multi Agent Syst.},
  volume       = {36},
  number       = {2},
  pages        = {41},
  year         = {2022},
  url          = {https://doi.org/10.1007/s10458-022-09575-5},
  doi          = {10.1007/S10458-022-09575-5},
  bibsource    = {dblp computer science bibliography, https://dblp.org}
}

@article{hsu2023safety,
  title={The safety filter: A unified view of safety-critical control in autonomous systems},
  author={Hsu, Kai-Chieh and Hu, Haimin and Fisac, Jaime F},
  journal={Annual Review of Control, Robotics, and Autonomous Systems},
  volume={7},
  year={2023},
  publisher={Annual Reviews}
}

@InProceedings{pmlr-v80-riedmiller18a,
  title = 	 {Learning by Playing Solving Sparse Reward Tasks from Scratch},
  author =       {Riedmiller, Martin and Hafner, Roland and Lampe, Thomas and Neunert, Michael and Degrave, Jonas and van de Wiele, Tom and Mnih, Vlad and Heess, Nicolas and Springenberg, Jost Tobias},
  booktitle = 	 {Proceedings of the 35th International Conference on Machine Learning},
  pages = 	 {4344--4353},
  year = 	 {2018},
  editor = 	 {Dy, Jennifer and Krause, Andreas},
  volume = 	 {80},
  series = 	 {Proceedings of Machine Learning Research},
  month = 	 {10--15 Jul},
  publisher =    {PMLR},
  url = 	 {https://proceedings.mlr.press/v80/riedmiller18a.html}
}

@article{chen2021decision,
  title={Decision transformer: Reinforcement learning via sequence modeling},
  author={Chen, Lili and Lu, Kevin and Rajeswaran, Aravind and Lee, Kimin and Grover, Aditya and Laskin, Misha and Abbeel, Pieter and Srinivas, Aravind and Mordatch, Igor},
  journal={Advances in neural information processing systems},
  volume={34},
  pages={15084--15097},
  year={2021}
}

@book{precup2000temporal,
  title={Temporal abstraction in reinforcement learning},
  author={Precup, Doina},
  year={2000},
  publisher={University of Massachusetts Amherst}
}

@article{huang2020closer,
  title={A closer look at invalid action masking in policy gradient algorithms. arXiv 2020},
  author={Huang, Shengyi and Onta{\~n}{\'o}n, Santiago},
  journal={arXiv preprint arXiv:2006.14171},
  year={2020}
}

@article{gu2022review,
  title={A review of safe reinforcement learning: Methods, theory and applications},
  author={Gu, Shangding and Yang, Long and Du, Yali and Chen, Guang and Walter, Florian and Wang, Jun and Knoll, Alois},
  journal={arXiv preprint arXiv:2205.10330},
  year={2022}
}

@article{laroche2017multi,
  title={Multi-advisor reinforcement learning},
  author={Laroche, Romain and Fatemi, Mehdi and Romoff, Joshua and van Seijen, Harm},
  journal={arXiv preprint arXiv:1704.00756},
  year={2017}
}

@ARTICLE{Thananjeyan2021recovery,
  author={Thananjeyan, Brijen and Balakrishna, Ashwin and Nair, Suraj and Luo, Michael and Srinivasan, Krishnan and Hwang, Minho and Gonzalez, Joseph E. and Ibarz, Julian and Finn, Chelsea and Goldberg, Ken},
  journal={IEEE Robotics and Automation Letters}, 
  title={Recovery RL: Safe Reinforcement Learning With Learned Recovery Zones}, 
  year={2021},
  volume={6},
  number={3},
  pages={4915-4922},
  doi={10.1109/LRA.2021.3070252}
}

\end{document}